\documentclass[10pt,twocolumn,letterpaper]{article}

\usepackage[pagenumbers]{wacv} 

\usepackage{graphicx}
\usepackage{amsmath}
\usepackage{amssymb}
\usepackage{booktabs}

\usepackage{array}  
\newcolumntype{P}[1]{>{\centering\arraybackslash}p{#1}}

\usepackage{multirow}

\usepackage[pagebackref,breaklinks,colorlinks]{hyperref}

\usepackage[capitalize]{cleveref}
\crefname{section}{Sec.}{Secs.}
\Crefname{section}{Section}{Sections}
\Crefname{table}{Table}{Tables}
\crefname{table}{Tab.}{Tabs.}

\def\wacvPaperID{2282} 
\def\confName{WACV}
\def\confYear{2025}

\begin{document}

\title{Continuous Spatio-Temporal Memory Networks for 4D Cardiac Cine MRI Segmentation}


\author{Meng Ye$^{1}$, Bingyu Xin$^{1}$, 
Leon Axel$^{2}$, Dimitris Metaxas$^{1}$\\
$_{}^{1}\textrm{}$Rutgers University, $_{}^{2}\textrm{}$New York University School of Medicine\\
{\tt\small \{my389, bx64, dnm\}@cs.rutgers.edu }
}


\maketitle

\begin{abstract}
   Current cardiac cine magnetic resonance image (cMR) studies focus on the end diastole (ED) and end systole (ES) phases, while ignoring the abundant temporal information in the whole image sequence. This is because whole sequence segmentation is currently a tedious process and inaccurate. Conventional whole sequence segmentation approaches first estimate the motion field between frames, which is then used to propagate the mask along the temporal axis. However, the mask propagation results could be prone to error, especially for the basal and apex slices, where through-plane motion leads to significant morphology and structural change during the cardiac cycle. Inspired by recent advances in video object segmentation (VOS), based on spatio-temporal memory (STM) networks, we propose a continuous STM (CSTM) network for semi-supervised whole heart and whole sequence cMR segmentation. Our CSTM network takes full advantage of the spatial, scale, temporal and through-plane continuity prior of the underlying heart anatomy structures, to achieve accurate and fast 4D segmentation. Results of extensive experiments across multiple cMR datasets show that our method can improve the 4D cMR segmentation performance, especially for the hard-to-segment regions. Project page is at \url{https://github.com/DeepTag/CSTM}.
\end{abstract}

\begin{figure}[t]
\begin{center}
\includegraphics[width=1.0\linewidth]{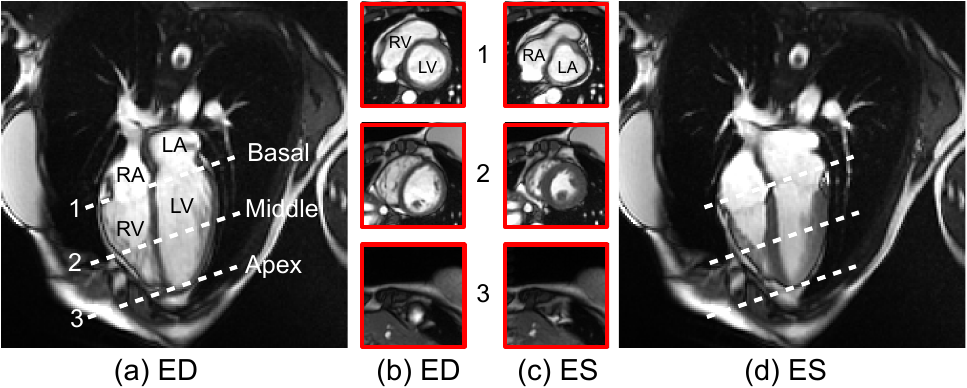}
\end{center}
   \caption{Cardiac cine magnetic resonance (cMR) images. (a) and (d) show the long-axis 4 chamber views; (b) and (c) show the short-axis views at/near (1) basal, (2) middle and (3) apex region. While (a) and (b) show the images at the end diastole (ED) phase, (c) and (d) show the images at the end systole (ES) phase. Through-plane motion of the heart causes in-plane structural change, which can be observed in (b) and (c), especially for the basal and apex slices. The red box in (b) and (c) shows the area of/nearby the heart ventricles. Note that slice position 1 corresponds to the left ventricle (LV) and right ventricle (RV) at ED,  but to the left atrium (LA) and right atrium (RA) at ES;  slice position 3 has an intersection with the LV apex at ED, but not at ES.}
\label{fig1}
\end{figure}

\section{Introduction}
\label{sec:intro}
Cardiac cine magnetic resonance (cMR) imaging is the gold standard for heart function evaluation~\cite{amzulescu2019Myocardial}. Limited by the slow MR imaging speed, current clinical scan protocols are 2D-based. As shown in Fig.~\ref{fig1}, to cover the whole heart region, we usually acquire stacks of 2D short-axis image sequences along the long-axis of the heart. While complete motion information of the heart is present in the cine sequence, current quantitative function assessments of cMR images, e.g., the stroke volume (SV) and the ejection fraction (EF), just focus on the end diastole (ED) and end systole (ES) phases, which could lead to potential inaccurate heart disease diagnosis and less effective treatment~\cite{wu2014evaluation}. Failure to analyze the complete image sequence is due in large part to the lack of reliable and efficient methods for whole sequence segmentation~\cite{lee2009automatic, barbaroux2023automated}. Conventional whole sequence cMR segmentation relies on a two-step approach: in-plane heart motion between frames is first estimated~\cite{ye2021deeptag}, and then is used to propagate the 2D segmentation mask through the cardiac cycle~\cite{suinesiaputra2014collaborative, ye2023sequencemorph}. While intuitive and simple, mask propagation results from this approach are prone to error, especially for the basal and apex slices, where through-plane motion of the heart leads to significant in-plane morphology and structural changes over time, as can be observed in Fig.~\ref{fig1} (b) and (c). Recent single-frame segmentation models based on deep convolutional neural networks~\cite{ronneberger2015u} and vision transformers~\cite{gao2021utnet} may also fail to properly segment the basal and apex regions~\cite{campello2021multi}. One reason leading to such failure is that both  training and inference of those models are typically based on a single frame, without explicitly using the temporal coherence in the image sequence. 

More recently, the task of mask propagation in an video has been formulated as a semi-supervised video object segmentation (VOS) problem~\cite{perazzi2016benchmark, zhou2022survey}. The idea of a spatio-temporal memory (STM) network~\cite{oh2019video} has greatly advanced VOS performance for natural scenes. The success of STM lies in  fully exploiting existing and intermediate segmentation results with corresponding frames, the \textbf{memory}, in an video, and the dense matching between the \textbf{query}, the current frame to be segmented, and the memory, by using the \textbf{attention} mechanism~\cite{vaswani2017attention}. Following this idea, a series of methods have been developed to improve the dense matching accuracy between query and memory~\cite{cheng2021rethinking}, or to construct a more efficient memory~\cite{cheng2022xmem}. For medical image segmentation,  recently introduced interactive methods~\cite{liu2022isegformer} formulate the mask propagation process from an annotated slice to remaining slices as being the same as a VOS task~\cite{zhou2023volumetric}. Although memory-based networks improve the mask propagation performance during interactive medical image segmentation, no corresponding mask propagation methods have been proposed, which are dedicated to volumetric images.


This work is dedicated to the semi-supervised 4D cardiac cine MRI VOS problem. We assume that the segmentation mask of the first frame (typically ED phase) in the middle wall region exists, and we design a continuous spatio-temporal memory (CSTM) network to propagate the mask to remaining frames and slices, so that we can obtain whole heart and whole sequence cMR segmentation. We are inspired by the 4D spatio-temporal continuity of the underlying heart anatomical structures in the image sequences to develop our CSTM network, which takes full advantage of the \textit{spatial, scale, temporal and through-plane continuity} of an image volume over time, to achieve accurate, fast, whole heart and whole sequence cMR segmentation.
Our main contributions are summarized as following: (1) We proposed patch-level matching to filter out noisy memory matching and to efficiently use multi-level memory features. (2) We proposed 4D inference strategy which greatly benefits the 4D cMR segmentation. (3) We performed extensive experiments on three cMR datasets to validate the efficacy of our method. 

\section{Related Work}
\subsection{Cardiac Cine MRI and Its Segmentation}
Although MR imaging can directly provide motion information on the beating heart by cine imaging, its inherently slow imaging speed makes current clinical cMR scan protocols 2D-based. By gradually shifting the cine imaging slice position from the apex to the base of the heart, we obtain the whole heart dynamic imaging data, in the form of a stack of 2D cMR image sequences. Deep convolutional neural networks~\cite{ronneberger2015u, khened2019fully, isensee2021nnu} and attention-based vision transformer networks~\cite{gao2021utnet, gao2022data, zhou2023nnformer} have been broadly applied to the automatic segmentation of cMR images. Several benchmarks, e.g., ACDC~\cite{bernard2018deep}, MnM~\cite{campello2021multi}, and MnM-2~\cite{martin2023deep}, have greatly boosted the associated work. However, most previous cMR segmentation models are based on a single frame at a given slice, without the ability to exploit temporal coherence between frames, which leads to inaccurate segmentation of the basal and apex slices. 

Conventional whole sequence cMR segmentation approaches rely on in-plane motion estimation of the moving heart~\cite{li2010line, suinesiaputra2014collaborative, ye2023sequencemorph}. However, the through-plane motion of the heart cannot be accounted into the in-plane motion estimate alone. Thus, erroneous mask propagation results are often found in basal and apex regions, where through-plane motion significantly changes the in-plane heart structures seen in the fixed imaging plane over time. 
The mask propagation between slices along the long axis ($z$-axis) is challenging, because of the large background variation, which leads to inaccurate deformation field estimation.
Therefore, it is common in conventional approaches that all $z$-axis slices at the starting time point have been segmented to achieve whole heart and whole sequence segmentation. 
We utilize not only the temporal coherence in an image sequence, but also the spatial coherence of the underlying heart, which spans the full 3D space, to achieve better whole heart and whole sequence cMR segmentation, while only a single initial slice segmentation is required.

\begin{figure*}[t]
\begin{center}
\includegraphics[width=0.7\linewidth]{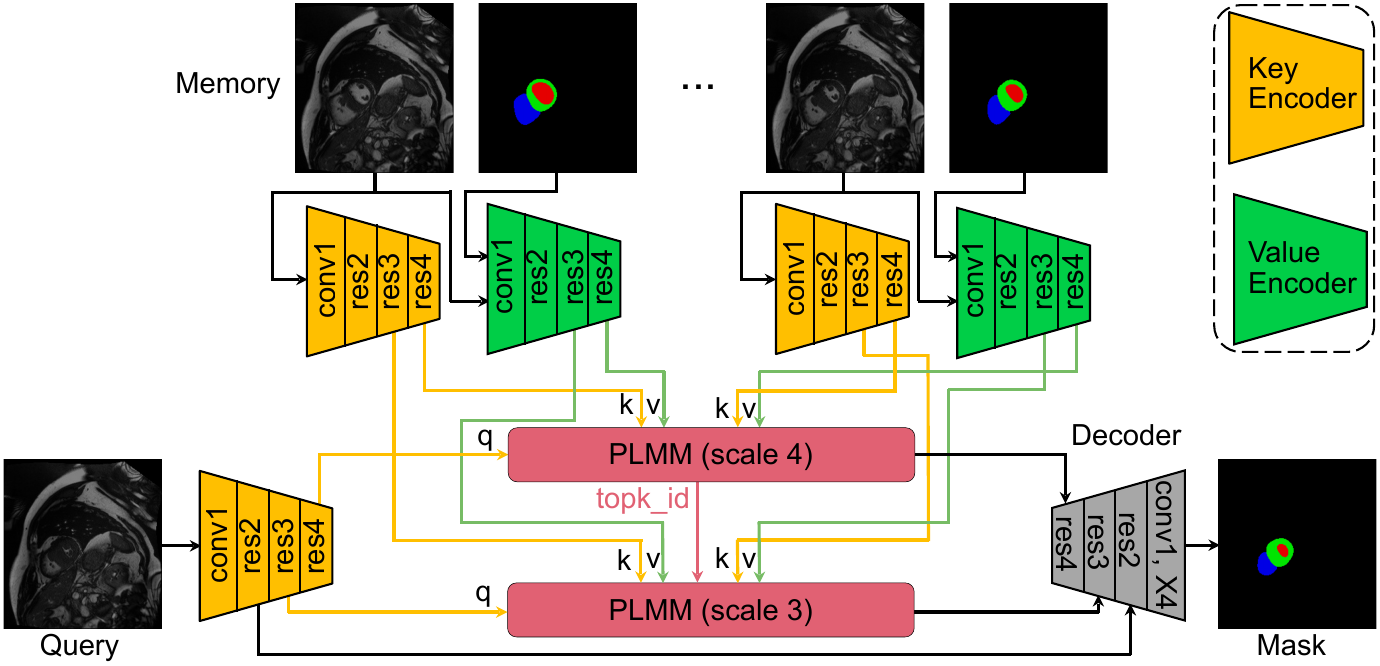}
\end{center}
   \caption{Architecture of CSTM. Both key and value encoders are based on ResNet. The key encoder takes a query/memory frame as input and outputs multi-scale query/memory key features. The value encoder inputs a memory frame with the corresponding segmentation mask and outputs multi-scale value features. We perform patch-level memory matching (PLMM) at scale 3 and 4 to read out memory values, which are fed into the decoder to output the segmentation mask of the query frame.}
\label{fig2}
\end{figure*}

\subsection{Memory-Based Video Object Segmentation}
While there are various approaches for video object segmentation (VOS)~\cite{perazzi2016benchmark, zhou2022survey}, we focus on a semi-supervised one. Object appearances changing over time, occlusions, similar surrounding distractors, and imaging noise make accurate mask propagation in a video challenging. Early VOS methods were based on online propagation~\cite{hu2017maskrnn, perazzi2017learning, li2018video
}, object detection~\cite{caelles2017one, bao2018cnn, chen2018blazingly
}, and hybrid networks~\cite{oh2018fast, yang2018efficient}. Spatio-temporal memory (STM) network~\cite{oh2019video} is a pioneering work which has greatly advanced VOS performance. Following this work, new methods have been developed to obtain more accurate matching between the query and the memory~\cite{cheng2021rethinking, wang2023look, zhang2023boosting, sun2023alignment, cheng2024putting}, and to construct more efficient memory~\cite{cheng2022xmem}. For multi-object VOS, multi-object association is achieved through an identification mechanism~\cite{yang2021associating, yang2022decoupling}. Recently, the vision transformer (ViT) network has been used to jointly extract features from query and memory frames and to model the interaction between query-memory features, without explicit query-memory matching~\cite{wu2023scalable, zhang2023joint}. To train such ViTs for VOS, large scale pretraining, e.g., MAE~\cite{he2022masked}, is necessary. Despite achieving high mask propagation accuracy, the inference speed of those methods is relatively slow, because of the quadratic computational complexity of self-attention in ViTs.

Memory-based VOS has been applied to interactive medical image segmentation~\cite{liu2022isegformer, zhou2023volumetric}. In a volumetric image, a center slice of an organ is first segmented, and the mask is then propagated bi-directionally to remaining slices, given that the in-plane organ appearance is similar to each other and changes slowly across slices. However, no mask propagation method has been proposed to be fully adaptive to properties of dynamic volumetric medical images. Therefore, we designed a novel mask propagation method, accounting for the 4D continuity in cMR image sequences, to achieve accurate and efficient segmentation.

\section{Methods}
The architecture of CSTM is shown in Fig.~\ref{fig2}. The design of CSTM is based on STCN~\cite{cheng2021rethinking},  which consists of a key encoder, a value encoder and a mask decoder. The key encoder is a Siamese structure~\cite{bromley1993signature}, which is shared by the query and memory frames. Thus, the key encoder can map the query frame and memory frames into the same key feature space, where correspondence can be easily established by comparing the affinity scores. Formally, given a query frame and $T$ memory frames, the key encoder maps the images into the following key features: query key $\mathbf{q}\in \mathbb{R}^{C_{k}\times HW}$ and memory key $\mathbf{k}\in \mathbb{R}^{C_{k}\times THW}$.  
Differently from STCN, we design a patch-level memory matching (PLMM) module to determine the query-memory matching weights\footnote[1]{As we will show next, the matching weight matrix for query-memory patches takes a different form.} $\mathbf{w}\in \mathbb{R}^{THW\times HW}$. The value encoder encodes memory frames and their corresponding segmentation masks as memory values: $\mathbf{v}\in \mathbb{R}^{C_{v}\times THW}$. With the query-memory matching weights, we can read out memory values: $\mathbf{v}_{ro} =\mathbf{v}\mathbf{w}$, where $\mathbf{v}_{ro}\in \mathbb{R}^{C_{v}\times HW}$. The retrieved memory values $\mathbf{v}_{ro}$ are then fed into the mask decoder to predict the segmentation mask of the query frame. We perform query-memory matching and memory value read out at both the coarse and fine scales (scale 3 and 4), which is another difference from STCN. Below, we give details of CSTM.

\begin{figure}[t]
\begin{center}
\includegraphics[width=1.0\linewidth]{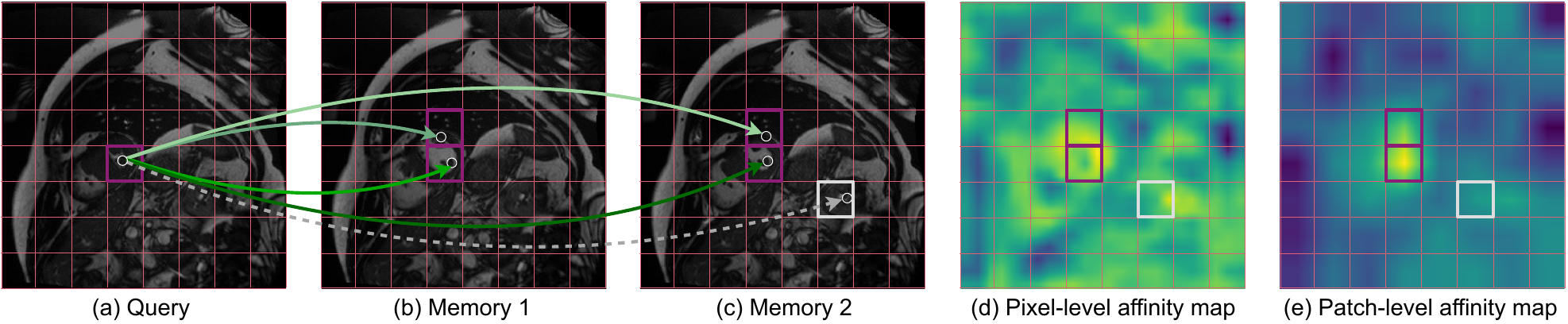}
\end{center}
   \caption{An illustration of patch-level memory matching (PLMM). We first divide an image or feature map into patches, then we match each query patch in (a) with top-K memory patches in (b) and (c). Finally, we perform dense pixel-level matching between the query patch and the top-K memory patches. In (d), we show the pixel-level affinity map of the query pixel in (a) with memory 2 in (c). In (e), we show the patch-level affinity map of the query patch in (a) with memory 2 in (c). PLMM can efficiently filter out noisy matches, e.g., the dashed line between the \textit{left ventricle} area (purple) and the \textit{stomach} area (gray), by leveraging the local spatial continuity prior in an image.}
\label{fig3}
\end{figure}

\subsection{Patch-Level Memory Matching}
Our patch-level memory matching (PLMM) module makes use of the local spatial continuity in an image to efficiently filter out noisy query-memory matches. Previous methods perform dense query-memory matching at pixel level, which could result in false matching because of noise in the key feature maps. For cMR images, due to the lack of image contrast, the key feature noise could be even more worse compared with natural images.
For example, as shown in Fig.~\ref{fig3}, pixels of the \textit{left ventricle} area (purple) in the query frame are often matched with pixels of the \textit{stomach} area (gray) in the memory frame, because of their similar appearance. However, we can easily observe that a local image patch often clusters with pixels of the same semantic region, i.e., the local spatial continuity prior in an image. Therefore, we first perform patch-level query-memory matching to filter out noisy memory patches, then we perform pixel-level matching within the matched query and memory patches. In this way, we can not only establish robust query-memory matching, but also greatly reduce the computational complexity of matching, as shown in the Supplementary Material.

Formally, we first divide the key feature maps and memory value maps into \textit{overlapping} patches:
$\mathbf{q}\in \mathbb{R}^{C_{k}\times NH_{p}W_{p}}$, $\mathbf{k}\in \mathbb{R}^{C_{k}\times TNH_{p}W_{p}}$, $\mathbf{v}\in \mathbb{R}^{C_{v}\times TNH_{p}W_{p}}$, where $N$ is the total number of patches generated in a key feature or memory value map, $H_{p}$ and $W_{p}$ are the height and width of the feature patch, respectively. For simplicity, we set $H_{p}=W_{p}=P$. To ensure each pixel can be located in the center of a patch, we set the overlapping size as $P/2$. In our implementation, we use the \texttt{unfold} operation in PyTorch to finish this patch dividing process.

Then, we compute the patch-level affinity scores:
\begin{equation}
\mathbf{A}_{patch}=d(\mathbf{q}_{patch}, \mathbf{k}_{patch}), 
\end{equation}
where $\mathbf{q}_{patch}\in \mathbb{R}^{N\times C_{k}H_{p}W_{p}}$ and $\mathbf{k}_{patch}\in \mathbb{R}^{TN\times C_{k}H_{p}W_{p}}$ are the flattened query and key patches, $d(\cdot, \cdot)$ is the $l_{2}$ distance following~\cite{cheng2021rethinking}, and $\mathbf{A}\in \mathbb{R}^{N\times TN}$.
With $\mathbf{A}_{patch}$, we can search the top-K matched $K$ memory patches for each query patch: $\mathbf{k}^{topk}_{patch}=\mathbf{k}[topk_{-}id]$, where $\mathbf{k}^{topk}_{patch}\in \mathbb{R}^{N\times C_{k}KH_{p}W_{p}}$, $topk_{-}id$ is a mapping which selects out the top-K memory patches from the $TN$ memory patches for each query patch. In the same way, we can select out the corresponding top-K matched memory value patches: $\mathbf{v}^{topk}_{patch}=\mathbf{v}[topk_{-}id]$, where $\mathbf{v}^{topk}_{patch}\in \mathbb{R}^{N\times C_{v}KH_{p}W_{p}}$. As shown in Fig.~\ref{fig3}, the top-K filtering can eliminate false patch matching (gray dashed line) with the help of a local spatial continuity prior.

Next, we compute the pixel-level query-memory matching weights within the matched query-memory patches, in the form of \texttt{softmax}: 
\begin{equation}
    \mathbf{w}_{patch}[i, j]=\frac{exp(d(\mathbf{q}_{patch}[i], \mathbf{k}^{topk}_{patch}[j]))}{\sum_{j}exp(d(\mathbf{q}_{patch}[i], \mathbf{k}^{topk}_{patch}[j]))},
\label{eq2}
\end{equation}
where $d(\cdot, \cdot )$ is the $l_{2}$ distance, $\mathbf{w}_{patch}\in \mathbb{R}^{NH_{p}W_{p}\times KH_{p}W_{p}}$, $i\in [1, H_{p}W_{p}]$, $j\in [1, KH_{p}W_{p}]$. We show an example in Fig.~\ref{fig3}. For a query pixel (white circle) in the query patch, the pixel-level dense matching in Eq.~(\ref{eq2}) can determine the matching weights with pixels (white circles) in the top-4 matched memory patches. Although the top-K filtering in the patch matching process is not differentiable, the \texttt{softmax} operation in Eq.~(\ref{eq2}) is differentiable. Thus, by training the CSTM network, it can learn to assign proper matching weights (different depth of green lines) in $\mathbf{w}_{patch}$. With $\mathbf{w}_{patch}$ and  $\mathbf{v}^{topk}_{patch}$, we read out the memory values\footnote[2]{In practice, we properly permute the tensors to calculate the memory readout values; however, it takes the form of the tensor product shown in Eq.~(\ref{eq3}). }:
\begin{equation}
    \mathbf{v}_{ro,\:patch} =\mathbf{v}^{topk}_{patch}\mathbf{w}_{patch},
\label{eq3}
\end{equation}
where $\mathbf{v}_{ro,\:patch}\in \mathbb{R}^{C_{v}\times NH_{p}W_{p}}$.


Lastly, we use the \texttt{fold} operation to combine the array of $\mathbf{v}_{ro,\:patch}$ into a memory value map:
\begin{equation}
    \mathbf{v}_{ro} = fold(\mathbf{v}_{ro,\:patch}),
\end{equation}
where $\mathbf{v}_{ro}\in \mathbb{R}^{C_{v}\times HW}$. The overlapped areas in $\mathbf{v}_{ro,\:patch}$ are averaged, in order to alleviate the blocking effect in the final output of $\mathbf{v}_{ro}$.

\begin{figure}[t]
\begin{center}
\includegraphics[width=1.0\linewidth]{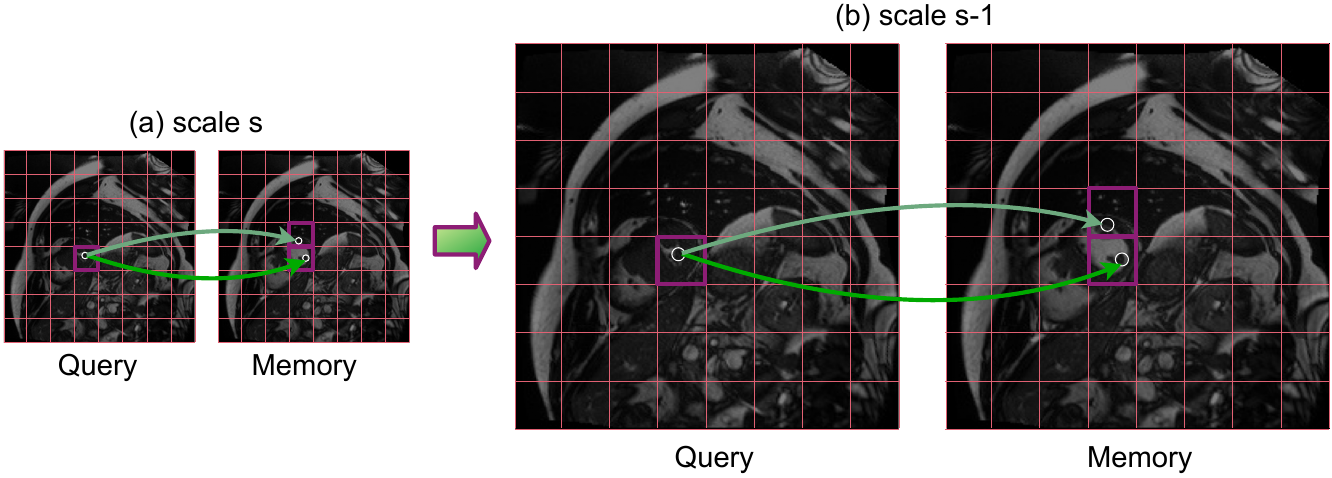}
\end{center}
   \caption{Multi-scale memory matching by leveraging the scale continuity prior in a feature pyramid. The feature map size at scale $s$ is 1/4 of that at scale $s-1$. We set the patch size at scale $s-1$ as 4$\times$ of that at scale $s$. After performing patch matching at scale $s$, we directly copy the $topk_{-}id$ to scale $s-1$. The scale continuity prior can ensure the patch matching accuracy at the coarser scale $s-1$ by passing the matching results across scales.}
\label{fig4}
\end{figure}

\subsection{Multi-Scale Memory Matching}
We perform memory matching at both low and high resolution scales. Although accurate matching can be achieved at the coarse scale with higher level semantic features, fine-grained features within the high resolution scale can result in more accurate segmentation masks~\cite{seong2021hierarchical}. The prohibitively heavy computation for dense self-attention matching at the fine scale prevents multi-scale memory matching in previous work~\cite{oh2019video, cheng2021rethinking}. In addition to the PLMM design, we leverage the scale continuity prior within a feature pyramid to achieve accurate and fast multi-scale memory matching.

\begin{figure*}[t]
\begin{center}
\includegraphics[width=0.7\linewidth]{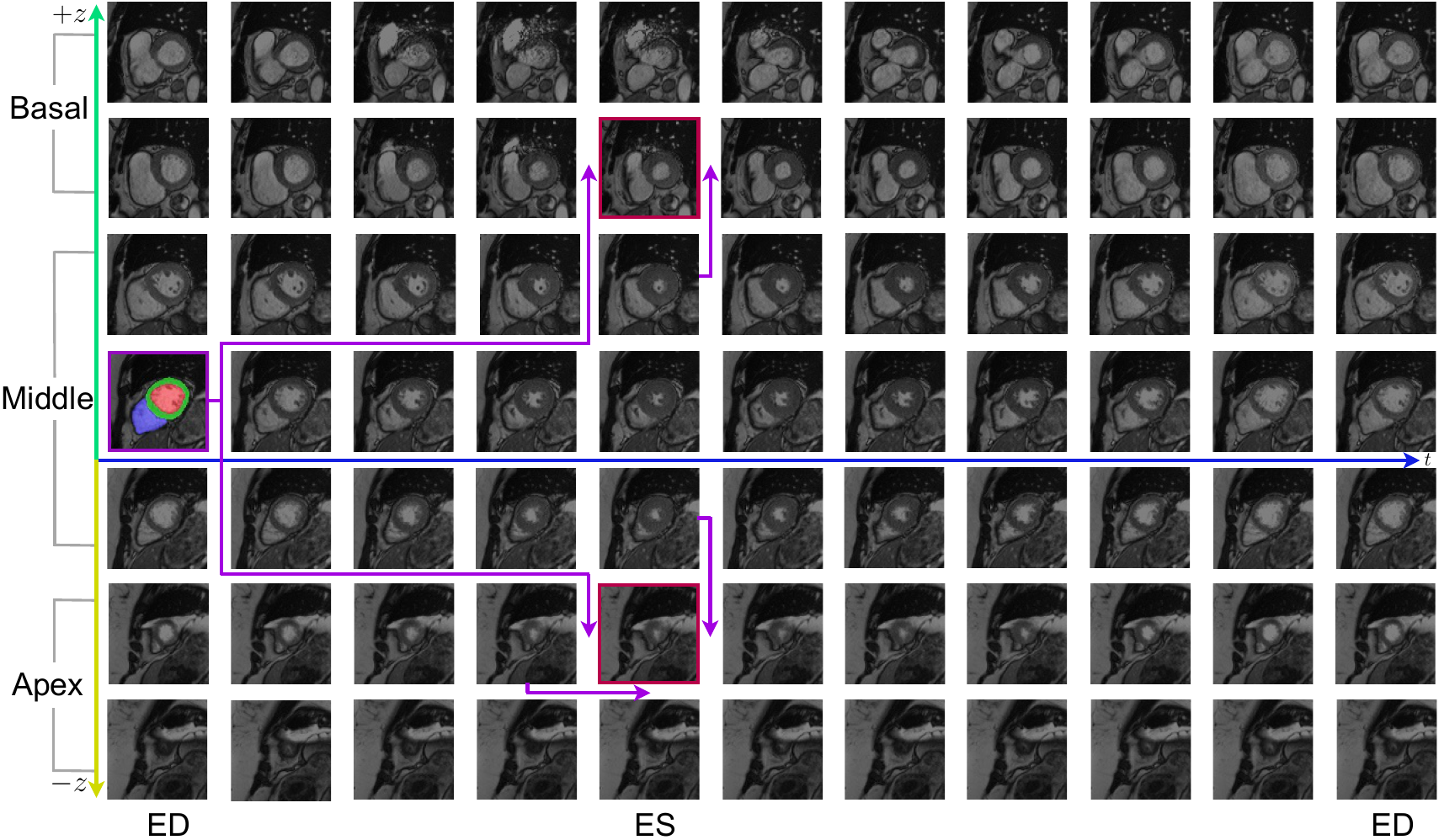}
\end{center}
   \caption{An illustration of the inference strategy of CSTM. The annotated frame is shown in the purple box ($t_{z0}=0$), in which red is the LV, green is the myocardium wall (Myo), and blue is the RV. We propagate the mask first along the temporal $t$-axis (blue line); then along the $z$-axis (green and yellow lines). For each query frame $t_{z}=\tau$ (red box) in the basal or middle region, we use the memory at $t_{z0}=0$ and $t_{z-1}=\tau$ for spatio-temporal matching. For each query frame $t_{z}=\tau$ in the apex region, we use the memory at $t_{z0}=0$, $t_{z+1}=\tau$ and $t_{z}=\tau-1$, for spatio-temporal matching.}
\label{fig5}
\end{figure*}

As shown in Fig.~\ref{fig4}, the feature map size at scale $s$ is 1/4 of that at scale $s-1$. We set the patch size at scale $s-1$ as 4$\times$ of that at scale $s$: ${H}'_{p}={W}'_{p}=2P$. We first perform patch matching at scale $s$, then we directly assign the $topk_{-}id$ to scale $s-1$ to match the query-memory patches. At each scale, since the local spatial continuity prior holds, this direct crossing scale matching results passing should be guaranteed accurate. After we match the patches at scale $s-1$, we compute the pixel-level matching weights and read out the memory values in the same manner as at scale $s$.

Our method is different from the top-K guided memory matching method in~\cite{seong2021hierarchical}. In their method, dense memory matching is first performed at the coarse scale; then, the top-K matching guidance for each pixel is mapped to a finer scale by using the scale continuity. The dense matching at the coarse scale can result in false matching because of noisy features, which will be propagated to the finer scale. We believe this why their model tends to converge to a sub-optimum without the dropout layer at the finer scale. In contrast, we never observe the sub-optimal convergence results with our multi-scale memory matching model.

\begin{table*}[t]
\begin{center}
\begin{tabular}{|P{1.4cm}|P{1.2cm}|P{1.2cm}|P{1.2cm}|P{1.2cm}|P{1.2cm}|P{1.2cm}|P{1.2cm}|P{1.2cm}|}
\hline
\multirow{2}{*}{Method} &\multicolumn{4}{c|}{Dice $\uparrow$} &\multicolumn{4}{c|}{HD ($mm$) $\downarrow$} \\
\cline{2-9}
 & LV & Myo& RV & Avg & LV & Myo& RV & Avg\\
\hline
\multicolumn{9}{|c|}{Basal}\\
\hline
STM &$0.944$ &$0.899$ &$0.929$ &$0.924$ &$1.349$ &$2.006$ &$3.250$ &$2.201$ \\
HMMN &$0.941$ &$0.892$ &$0.930$ &$0.921$ &$1.497$ &$2.267$ &$2.988$ &$2.251$ \\
STCN &$0.940$ &$0.897$ &$0.910$ &$0.916$ &$1.476$ &$2.019$ &$3.475$ &$2.323$ \\
XMem &$0.941$ &$0.895$ &$0.930$ &$0.922$ &$1.385$ &$2.057$ &$3.211$ &$2.218$ \\
Ours &$\mathbf{0.950}$ &$\mathbf{0.905}$ &$\mathbf{0.938}$ &$\mathbf{0.931}$ &$\mathbf{1.249}$ &$\mathbf{1.876}$ &$\mathbf{2.836}$ &$\mathbf{1.987}$ \\
\hline
\multicolumn{9}{|c|}{Middle}\\
\hline
STM &$0.951$ &$0.906$ &$0.910$ &$0.923$ &$1.415$ &$1.736$ &$2.552$ &$1.901$ \\
HMMN &$0.953$ &$0.907$ &$0.914$ &$0.924$ &$1.388$ &$1.653$ &$2.476$ &$1.839$ \\
STCN &$0.956$ &$0.914$ &$0.923$ &$0.931$ &$1.212$ &$1.522$ &$2.101$ &$1.612$ \\
XMem &$0.954$ &$0.910$ &$0.922$ &$0.929$ &$1.210$ &$1.651$ &$2.077$ &$1.646$ \\
Ours &$\mathbf{0.956}$ &$\mathbf{0.916}$ &$\mathbf{0.924}$ &$\mathbf{0.932}$ &$\mathbf{1.172}$ &$\mathbf{1.508}$ &$\mathbf{1.926}$ &$\mathbf{1.535}$ \\
\hline
\multicolumn{9}{|c|}{Apex}\\
\hline
STM &$0.866$ &$0.813$ &$0.836$ &$0.838$ &$2.685$ &$3.340$ &$3.452$ &$3.159$ \\
HMMN &$0.861$ &$0.814$ &$0.835$ &$0.836$ &$2.890$ &$3.590$ &$3.317$ &$3.266$ \\
STCN &$0.892$ &$0.857$ &$0.872$ &$0.874$ &$2.029$ &$2.306$ &$2.524$ &$2.286$ \\
XMem &$\mathbf{0.895}$ &$\mathbf{0.865}$ &$\mathbf{0.891}$ &$\mathbf{0.884}$ &$\mathbf{1.921}$ &$\mathbf{2.144}$ &$\mathbf{2.119}$ &$\mathbf{2.061}$ \\
Ours &$0.891$ &$0.861$ &$0.875$ &$0.876$ &$2.005$ &$2.472$ &$2.391$ &$2.289$ \\
\hline
\multicolumn{9}{|c|}{Whole Heart}\\
\hline
STM &$0.921$ &$0.873$ &$0.892$ &$0.895$ &$1.816$ &$2.360$ &$3.085$ &$2.420$ \\
HMMN &$0.918$ &$0.871$ &$0.893$ &$0.894$ &$1.925$ &$2.503$ &$2.927$ &$2.452$ \\
STCN &$0.929$ &$0.889$ &$0.902$ &$0.907$ &$1.572$ &$\mathbf{1.949}$ &$2.700$ &$2.074$ \\
XMem &$0.930$ &$0.890$ &$\mathbf{0.914}$ &$0.911$ &$1.506$ &$1.951$ &$2.469$ &$1.975$ \\
Ours &$\mathbf{0.932}$ &$\mathbf{0.894}$ &$0.913$ &$\mathbf{0.913}$ &$\mathbf{1.475}$ &$1.952$ &$\mathbf{2.384}$ &$\mathbf{1.937}$ \\
\hline
\end{tabular}
\end{center}
\caption{Comparisons between different 4D cMR segmentation methods on ACDC test set. We report the segmentation results at the basal, middle, apex and whole heart level, respectively.}
\label{table1}
\end{table*}

\begin{table*}[t]
\begin{center}
\begin{tabular}{|P{1.4cm}|P{1.2cm}|P{1.2cm}|P{1.2cm}|P{1.2cm}|P{1.2cm}|P{1.2cm}|P{1.2cm}|P{1.2cm}|}
\hline
\multirow{2}{*}{Method} &\multicolumn{4}{c|}{Dice $\uparrow$} &\multicolumn{4}{c|}{HD ($mm$) $\downarrow$} \\
\cline{2-9}
 & LV & Myo& RV & Avg & LV & Myo& RV & Avg\\
\hline
\multicolumn{9}{|c|}{Basal}\\
\hline
STM &$0.923$ &$0.886$ &$0.942$ &$0.917$ &$1.894$ &$2.405$ &$1.901$ &$2.067$ \\
HMMN &$0.919$ &$0.886$ &$0.939$ &$0.915$ &$2.059$ &$2.397$ &$1.903$ &$2.120$ \\
STCN &$0.883$ &$0.856$ &$0.918$ &$0.886$ &$2.698$ &$2.749$ &$2.381$ &$2.609$ \\
XMem &$0.905$ &$0.867$ &$0.927$ &$0.900$ &$2.169$ &$2.461$ &$2.116$ &$2.249$ \\
Ours &$\mathbf{0.923}$ &$\mathbf{0.889}$ &$\mathbf{0.945}$ &$\mathbf{0.919}$ &$\mathbf{1.814}$ &$\mathbf{2.127}$ &$\mathbf{1.736}$ &$\mathbf{1.892}$ \\
\hline
\multicolumn{9}{|c|}{Middle}\\
\hline
STM &$0.865$ &$0.800$ &$0.835$ &$0.834$ &$3.222$ &$3.420$ &$4.157$ &$3.600$ \\
HMMN &$0.865$ &$0.802$ &$0.834$ &$0.834$ &$3.231$ &$3.369$ &$4.160$ &$3.587$ \\
STCN &$0.871$ &$0.808$ &$0.842$ &$0.840$ &$3.110$ &$3.246$ &$4.073$ &$3.476$ \\
XMem &$0.852$ &$0.773$ &$0.822$ &$0.816$ &$3.562$ &$4.275$ &$4.506$ &$4.114$ \\
Ours &$\mathbf{0.873}$ &$\mathbf{0.815}$ &$\mathbf{0.844}$ &$\mathbf{0.844}$ &$\mathbf{3.046}$ &$\mathbf{3.179}$ &$\mathbf{3.924}$ &$\mathbf{3.383}$ \\
\hline
\multicolumn{9}{|c|}{Apex}\\
\hline
STM &$0.868$ &$0.816$ &$0.878$ &$0.854$ &$2.585$ &$3.088$ &$2.516$ &$2.730$ \\
HMMN &$0.868$ &$0.824$ &$0.872$ &$0.855$ &$2.514$ &$2.873$ &$2.578$ &$2.655$ \\
STCN &$0.873$ &$0.820$ &$0.871$ &$0.855$ &$2.475$ &$2.823$ &$2.563$ &$2.620$ \\
XMem &$0.853$ &$0.804$ &$0.874$ &$0.844$ &$2.773$ &$3.071$ &$2.551$ &$2.798$ \\
Ours &$\mathbf{0.875}$ &$\mathbf{0.830}$ &$\mathbf{0.881}$ &$\mathbf{0.862}$ &$\mathbf{2.320}$ &$\mathbf{2.806}$ &$\mathbf{2.423}$ &$\mathbf{2.516}$ \\
\hline
\multicolumn{9}{|c|}{Whole Heart}\\
\hline
STM &$0.885$ &$0.834$ &$0.885$ &$0.868$ &$2.567$ &$2.971$ &$2.858$ &$2.799$ \\
HMMN &$0.884$ &$0.837$ &$0.882$ &$0.868$ &$2.601$ &$2.880$ &$2.880$ &$2.787$ \\
STCN &$0.875$ &$0.828$ &$0.877$ &$0.860$ &$2.761$ &$2.939$ &$3.005$ &$2.902$ \\
XMem &$0.870$ &$0.815$ &$0.874$ &$0.853$ &$2.835$ &$3.269$ &$3.058$ &$3.054$ \\
Ours &$\mathbf{0.890}$ &$\mathbf{0.845}$ &$\mathbf{0.890}$ &$\mathbf{0.875}$ &$\mathbf{2.394}$ &$\mathbf{2.704}$ &$\mathbf{2.694}$ &$\mathbf{2.597}$ \\
\hline
\end{tabular}
\end{center}
\caption{Comparisons between different 4D cMR segmentation methods on MnM validation and test sets. We report the segmentation results at the basal, middle, apex and whole heart level, respectively.}
\label{table2}
\end{table*}

\subsection{4D Inference Strategy}
Considering the 4D nature of cMR sequences, the inference of CSTM is different from previous memory-based VOS models operating on natural scene videos. This is because we can make better use of the temporal and through-plane continuity prior in 4D cMR sequences to reduce computation, while maintaining the mask propagation accuracy. 

As shown in Fig.~\ref{fig5}, through-plane motion of the heart near the ES phase (shortening along the long axis) can cause in-plane content change for both basal and apex regions. For such regions, if we just propagate the mask along the temporal axis, the spatio-temporal continuity prior will be broken at some frames. However, for the middle region slices, the content change caused by through-plane heart motion during a cardiac cycle can be ignored. Therefore, we first propagate the mask along the temporal $t$-axis (blue line) within the middle slice 2D cMR sequence. We use the first frame and the previous frame, if available, as the memory ($T_{max}=2$).
Then at each phase, we propagate the mask along the $z$-axis (green and yellow lines). Although breath-holding inconsistency during 2D cine MR imaging can cause in-plane misalignment of the heart region~\cite{chang2021unsupervised}, the spatial continuity along the $z$-axis at each cardiac phase, i.e., through-plane continuity, always holds. In this way, we can make full use of the spatio-temporal continuity in the 4D cMR sequence for accurate query-memory matching. 

During mask propagation, for the basal and middle regions, we always take the memory at $t_{z0}=0$ and $t_{z-1}=\tau$ into the memory bank ($T_{max}=2$); for the apex region, we always take the memory at $t_{z0}=0$, $t_{z+1}=\tau$, and $t_{z}=\tau-1$, if available, into the memory bank ($T_{max}=3$).  We first perform multi-scale patch-level memory matching between the query frame ($t_{z}=\tau$) and the memory frames, to retrieve the top-K matched memory patches for each query patch. Then, for each pixel in a query patch, we perform pixel-level dense matching and read out the memory values with the computed matching weights.

\subsection{Implementation Details}
\textbf{Networks.}
Following previous work~\cite{cheng2021rethinking, cheng2022xmem}, we implemented the key encoder and value encoder with ResNet-50 and ResNet-18~\cite{he2016deep}, respectively. We removed the classification head and the last convolutional layer ($conv5_{-}x$) in each ResNet.
Thus, the coarsest scale of the resulting features is $4$ (stride=16). Since we used multi-scale memory matching, to reduce computational complexity, we set $C_{k}=32$ for both scale 3 and scale 4; $C_{v}=256$ for scale 3 and $C_{v}=512$ for scale 4.
The decoder in CSTM is close to that of STCN~\cite{cheng2021rethinking}. It has 3 residual connected layers with $2\times$ upsampling  and 1 convolutional layer. For each scale in scales 3 and 4, the memory read-out values are first concatenated with query key features along the channel dimension, and then fed into the corresponding layer in the decoder. The query key features at scale 2 are directly fed into the decoder, using skip-connections. The last layer of the decoder predicts a segmentation mask with stride 4, which is upsampled to the original image resolution, using $4\times$ bilinear interpolation. 

\textbf{Training and Testing.}
We curated a large 4D cMR dataset from three public 4D cMR datasets:  ACDC~\cite{bernard2018deep}, MnM~\cite{campello2021multi}, and MnM-2~\cite{martin2023deep}. Each 4D cMR set in these datasets only has mask annotations on the ED and ES phases. We trained and tested CSTM based on these sparsely annotated data. For each training iteration, we sampled three frames along the temporal axis (ED-ES-ED) or three spatially ordered slices along the $z$-axis as a training sample. We predicted the mask of the second frame with the first frame taken as the memory. The mask prediction, along with the first frame, will be used as the memory to predict the mask of the third frame. The maximum spatial sampling distance along the $z$-axis was set as 5. During patch-level memory matching, we set the patch size $P=6$, and $K=4$ for top-K patch matching. We used bootstrapped cross entropy loss following~\cite{cheng2021modular, cheng2021rethinking}. 
During inference, an A100 GPU was used with full floating point precision, for inference efficiency comparison.

\section{Experiments}
The experiments were conducted on the ACDC testing set and the MnM validation and testing sets. 
We compared our method with recent memory-based VOS methods: STM~\cite{oh2019video}, HMMN~\cite{seong2021hierarchical}, STCN~\cite{cheng2021rethinking}, and XMem~\cite{cheng2022xmem}. 
We trained these baseline models from scratch, following their original hyper-parameter settings. 
To evaluate the segmentation accuracy, we computed the Dice score~\cite{isensee2021nnu} and the $95$th quantile Hausdorff distance (HD) score~\cite{huttenlocher1993comparing} for the left ventricle (LV), myocardium wall (Myo), and right ventricle (RV), which are commonly used in medical image segmentation evaluation. 
To measure the inference efficiency, we use the frames per second (FPS) metric. 

\begin{table}
\begin{center}
\begin{tabular}{|P{1.0cm}|P{0.8cm}|P{1.0cm}|P{0.8cm}|P{0.9cm}|P{0.8cm}|}
\hline
 Method & STM & HMMN& STCN & XMem & Ours\\
\hline
FPS $\uparrow$ & $45.7$ & $16.0$ & $\mathbf{62.5}$  & $56.1$ & $44.9$ \\
\hline
\end{tabular}
\end{center}
\caption{Running time comparisons between different 4D cMR segmentation methods.}
\label{table3}
\end{table}

\subsection{Evaluation Results}

\begin{figure}
\begin{center}
\includegraphics[width=1.0\linewidth]{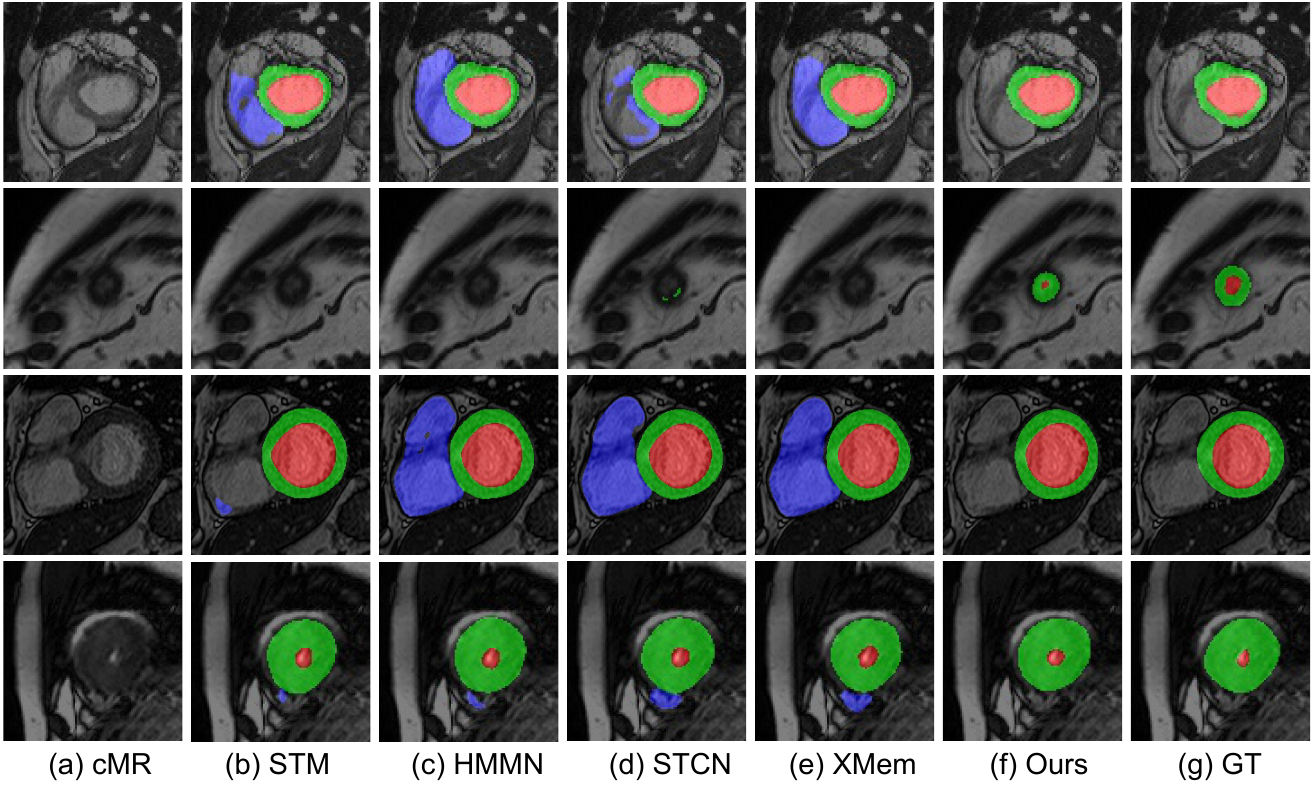}
\end{center}
   \caption{Qualitative comparison between different methods. The first two rows show the results on ACDC. The last two rows show the results on MnM. }
\label{fig6}
\end{figure}

\textbf{ACDC.} The results on ACDC is shown in Table~\ref{table1} and Fig.~\ref{fig6}. Our method outperforms all baseline methods on average. More specifically, it mainly improves the cMR segmentation performance for the basal and middle regions while it performs the second best for the apex regions. The through-plane motion affects the basal region more than the apex region. Thus, it is more challenging to segment the basal region. The results demonstrate that our method can improve the segmentation performance on the hard-to-segment instances near the base.

\textbf{MnM.} The results on MnM are shown in Table~\ref{table2} and Fig.~\ref{fig6}. Our method outperforms all baseline methods on average. For this dataset, it improves the cMR segmentation performance for all the basal, middle and apex regions. We note that, the MnM dataset is a highly heterogeneous cMR dataset, which consists of multi-center and multi-vendor datasets. In the test and validation dataset, it contains out-of-distribution subdatasets. Therefore, it is more challenging to segment the MnM dataset than the single-center ACDC dataset. This is why the segmentation performance drops for all the methods. While STCN and XMem perform better than STM and HMMN on the ACDC dataset, they perform worse on the more challenging MnM dataset. Our method, however, performs the best for both of the two datasets.

\textbf{Running Time.} During inference, we resized each input 2D image to ensure the shorter side has 384 pixels. Inference efficiency comparison is shown in Table~\ref{table3}. Our method ensures FPS $> 40$, which satisfies real-time inference efficiency requirement in clinical applications.
 
\begin{table}
\begin{center}
\begin{tabular}{|P{1.4cm}|P{1.2cm}|P{1.8cm}|P{1.2cm}|}
\hline
Scale & Dice $\uparrow$ & HD ($mm$) $\downarrow$ & FPS $\uparrow$\\
\hline
3  & $0.905$ & $2.197$ & $\mathbf{61.9}$  \\
\hline
4  & $0.909$ & $2.003$ & $61.5$  \\
\hline
3, 4  & $\mathbf{0.913}$ & $\mathbf{1.937}$ & $44.9$ \\
\hline
\end{tabular}
\end{center}
\caption{Ablation study of the feature scales used for query-memory matching.}
\label{table4}
\end{table}

\begin{table}
\begin{center}
\begin{tabular}{|P{1.4cm}|P{1.2cm}|P{1.8cm}|P{1.2cm}|}
\hline
 K & Dice $\uparrow$ & HD ($mm$) $\downarrow$ & FPS $\uparrow$\\
\hline
1  & $0.907$ & $2.080$  & $\mathbf{45.5}$ \\
\hline
2  & $0.911$ & $1.954$  & $45.2$ \\
\hline
4  & $\mathbf{0.913}$ & $\mathbf{1.937}$  & 44.9\\
\hline
6  & $0.913$ & $1.945$  & $42.1$\\
\hline
\end{tabular}
\end{center}
\caption{Ablation on the top-K query-memory patch matching.}
\label{table5}
\end{table}

\begin{table}
\begin{center}
\begin{tabular}{|P{1.4cm}|P{1.2cm}|P{1.8cm}|P{1.2cm}|}
\hline
 P & Dice $\uparrow$ & HD ($mm$) $\downarrow$ & FPS $\uparrow$\\
\hline
2  & $0.910$ & $2.019$  & $\mathbf{51.6}$\\
\hline
4  & $0.912$ & $1.943$  & $47.3$\\
\hline
6  & $\mathbf{0.913}$ & $\mathbf{1.937}$  & $44.9$\\
\hline
8  & $0.913$ & $1.971$  & $42.2$\\
\hline
\end{tabular}
\end{center}
\caption{Ablation on patch size used for query-memory matching.}
\label{table6}
\end{table}

\begin{table}
\begin{center}
\begin{tabular}{|P{1.4cm}|P{1.2cm}|P{1.8cm}|P{1.2cm}|}
\hline
Continuity & Dice $\uparrow$ & HD ($mm$) $\downarrow$ & FPS $\uparrow$\\
\hline
Spatial  & $0.910$ & $2.027$ & $44.2$  \\
\hline
Temporal  & $0.908$ & $2.038$ & $\mathbf{45.4}$  \\
\hline
Both  & $\mathbf{0.913}$ & $\mathbf{1.937}$  & $44.9$ \\
\hline
\end{tabular}
\end{center}
\caption{Ablation on spatial-temporal continuity for query-memory matching.}
\label{table7}
\end{table}

\begin{table}
\begin{center}
\begin{tabular}{|P{1.4cm}|P{1.2cm}|P{1.8cm}|P{1.2cm}|}
\hline
 $T_{max}$ & Dice $\uparrow$ & HD ($mm$) $\downarrow$ & FPS $\uparrow$\\
\hline
2  & $0.912$ & $1.945$ & $\mathbf{45.8}$  \\
\hline
3  & $0.913$ & $1.937$  & $44.9$ \\
\hline
5  & $0.914$ & $1.932$  & $43.5$ \\
\hline
10  & $0.915$ & $1.928$  & $40.9$ \\
\hline
15  & $\mathbf{0.915}$ & $\mathbf{1.926}$  & $34.1$ \\
\hline
\end{tabular}
\end{center}
\caption{Ablation on memory frames used for apex regions.}
\label{table8}
\end{table}

\subsection{Ablation Studies}
All of our ablation studies were conducted on the ACDC dataset as shown in the following:
(1) \textbf{Query-Memory Matching Scales.} In Table~\ref{table4}, we show that incorporation of both the coarse and fine scale query-memory matching can improve the segmentation performance.
(2) \textbf{Top-K Patch Matching.} In Table~\ref{table5}, we show that the Top-K number can affect the segmentation performance of our method. A smaller K would reduce useful memory information, while a larger K could introduce noise which leads to inaccurate query-memory matching.
(3) \textbf{Patch Size.} In Table~\ref{table6}, we show the effects of patch size on the segmentation performance of our method. Either a too small or too large patch cannot make use of the local spatial continuity and would introduce noisy query-memory matching results. 
(4) \textbf{Spatial-Temporal Continuity.} In our 4D inference strategy, we use both the spatial and temporal continuity to ensure accurate cMR segmentation. From Table~\ref{table7}, only using the temporal continuity, as in natural scene VOS applications, cannot obtain the optimal 4D cMR segmentation results.
(5) \textbf{Maximum Memory Frames for Apex Regions.} From Table~\ref{table8}, the apex region is small, which needs more memory information to accurately segment it.

\section{Conclusion}
In this work, we made full use of the spatial, scale, temporal and through-plane continuity prior in 4D cMR sequences, to ensure accurate whole heart and whole sequence segmentation. We performed extensive experiments and demonstrated that our method could improve the mask propagation performance on challenging-to-segment heart regions and out-of-distribution datasets. 

\section{Analysis of Computational Complexity} 
The computational complexity of pixel-level dense query-memory matching in STM~\cite{oh2019video}  or STCN~\cite{cheng2021rethinking} is $\mathcal{O}(TH^{2}W^{2})$. In our patch-level memory matching (PLMM), the main computational complexity consists of two parts: (1) $\mathcal{O}(TN^{2})$ for the computation of patch-level affinity scores and (2) $\mathcal{O}(NKH^{2}_{p}W^{2}_{p})$ for the computation of pixel-level query-memory matching weights within patches. Therefore, the overall computational complexity of PLMM is $\mathcal{O}(TN^{2}+NKH^{2}_{p}W^{2}_{p})$. Since $H_{p}\ll H$, $W_{p}\ll W$ and $K$ is usually small, the computational complexity of PLMM is greatly reduced, compared with the vanilla dense query-memory matching computational complexity. Furthermore, there is no \texttt{softmax} operation involved in the computation of patch-level affinity scores. Hence, our overall computational complexity is further reduced. 

\section{Dataset Details}
The 4D cMR dataset used in our work is curated from three public 4D cMR datasets:  ACDC~\cite{bernard2018deep}, MnM~\cite{campello2021multi}, and MnM-2~\cite{martin2023deep}. 
For ACDC, there are 100 cases in the training set and 50 cases in the testing set. For MnM, there are 175 cases in the training set, 34 cases in the validation set, and 136 cases in the testing set. For MnM-2, there are 160 cases in the training set and 40 cases in the validation set; the testing set is not released publicly. Apart from the validation set of MnM-2, mask annotations on the ED and ES phases of each 4D cMR set in these datasets are presented. We trained and tested CSTM based on these sparsely annotated cMR data. 
We combined the training sets in the three datasets as our own training set, which gives 435 4D cMR cases. 
And in the main text, we have reported the testing results on the ACDC testing set (50 cases) and the MnM validation and testing sets (170 cases), respectively.

All the three cMR datasets are of multiple cardiovascular pathologies and healthy volunteers. While ACDC is a single-center dataset, MnM and MnM-2 are both multi-center and multi-vendor datasets. Of particular, in the validation and testing datasets of MnM, there exists an out-of-distribution subdataset (center-5, MRI scanner manufacturer-Canon), which is not presented in the training dataset. Therefore, MnM is a more heterogeneous cMR dataset compared with ACDC.

\section{Training and Inference Details}
Each training sample consists of three frames, either sampled along the temporal axis (ED-ES-ED) or spatially ordered along the $z$-axis with a maximum spatial sampling distance of 5. 
The online data augmentation exactly follows the strategies used in the main training stage of STCN~\cite{cheng2021rethinking}. Basically, random horizontal flip, random resized cropping (of size 384), color jitter, random grayscale, and random affine transformation were included in the data augmentation. More details could be found in STCN~\cite{cheng2021rethinking}. 

We took at most two areas of the heart presenting on the first frame as segmentation targets. The key encoder (ResNet-50) and value encoder (ResNet-18) were pre-trained on ImageNet~\cite{deng2009imagenet}. The batch size was set as 4. Adam optimizer was used with default momentum $\beta_{1} = 0.9$, $\beta_{2} = 0.999$, a base learning rate of $10^{-5}$, and a $L_{2}$ weight decay of $10^{-7}$. The training was performed for 300K iterations.

During inference, we resized each input 2D cMR image to ensure the shorter side has a size of 384 pixels. The predicted segmentation mask was then resized to the original cMR image size.

\section{Training and Inference Details for Baseline Methods}
For baseline methods STM~\cite{oh2019video}, HMMN~\cite{seong2021hierarchical}, STCN~\cite{cheng2021rethinking}, and XMem~\cite{cheng2022xmem}, we follow the same training and inference schemes used for natural scene videos. For each 4D cMR data, we sequentially combined all 2D cMR sequences along the $z$-axis as a single long temporal sample, which was used for training or testing the baseline methods. Note that, each segment of 2D cMR sequence at a specific $z$ position covers a full cardiac cycle. Therefore, the temporal continuity is still preserved in the long temporal samples. 

We followed the original hyper-parameter settings to train these baseline models on the cMR dataset, with weights of their key and value encoders (ResNets) initialized from ImageNet~\cite{deng2009imagenet} pre-training.

During inference, all baseline methods took the first frame (at the middle myocardium wall level) with annotation masks as the memory and propagated the masks bi-directionally towards the basal or apex regions. 
For subsequent query frames, both STM and HMMN took every fifth frame as a memory frame, and the immediately previous frame as a temporary memory frame. 
STCN took every fifth query frame as a memory frame. 
XMem has three memory stores: sensory memory, working memory and long-term memory. 
The sensory memory was updated every query frame. 
The working memory was updated every fifth query frame following the First-In-First-Out (FIFO) approach to ensure the total number of memory frames $T_{max}\leq 5$. 
For most cases, the long-term memory was disabled since these temporal samples were not that long (usually less than 1K frames). 
When the long-term memory was enabled, we followed the long-term memory generation method proposed in XMem.

{\small
\bibliographystyle{ieee_fullname}
\bibliography{egbib}

\begin{thebibliography}{10}\itemsep=-1pt

\bibitem{amzulescu2019Myocardial}
Mihaela~Silvia Amzulescu, M De~Craene, H Langet, Agnes Pasquet, David Vancraeynest, Anne-Catherine Pouleur, Jean-Louis Vanoverschelde, and BL Gerber.
\newblock Myocardial strain imaging: review of general principles, validation, and sources of discrepancies.
\newblock {\em European Heart Journal-Cardiovascular Imaging}, 20(6):605--619, 2019.

\bibitem{bao2018cnn}
Linchao Bao, Baoyuan Wu, and Wei Liu.
\newblock Cnn in mrf: Video object segmentation via inference in a cnn-based higher-order spatio-temporal mrf.
\newblock In {\em Proceedings of the IEEE conference on computer vision and pattern recognition}, pages 5977--5986, 2018.

\bibitem{barbaroux2023automated}
Hugo Barbaroux, Karl~P Kunze, Radhouene Neji, Muhummad~Sohaib Nazir, Dudley~J Pennell, Sonia Nielles-Vallespin, Andrew~D Scott, and Alistair~A Young.
\newblock Automated segmentation of long and short axis dense cardiovascular magnetic resonance for myocardial strain analysis using spatio-temporal convolutional neural networks.
\newblock {\em Journal of Cardiovascular Magnetic Resonance}, 25(1):1--17, 2023.

\bibitem{bernard2018deep}
Olivier Bernard, Alain Lalande, Clement Zotti, Frederick Cervenansky, Xin Yang, Pheng-Ann Heng, Irem Cetin, Karim Lekadir, Oscar Camara, Miguel Angel~Gonzalez Ballester, et~al.
\newblock Deep learning techniques for automatic mri cardiac multi-structures segmentation and diagnosis: is the problem solved?
\newblock {\em IEEE transactions on medical imaging}, 37(11):2514--2525, 2018.

\bibitem{bromley1993signature}
Jane Bromley, Isabelle Guyon, Yann LeCun, Eduard S{\"a}ckinger, and Roopak Shah.
\newblock Signature verification using a" siamese" time delay neural network.
\newblock {\em Advances in neural information processing systems}, 6, 1993.

\bibitem{caelles2017one}
Sergi Caelles, Kevis-Kokitsi Maninis, Jordi Pont-Tuset, Laura Leal-Taix{\'e}, Daniel Cremers, and Luc Van~Gool.
\newblock One-shot video object segmentation.
\newblock In {\em Proceedings of the IEEE conference on computer vision and pattern recognition}, pages 221--230, 2017.

\bibitem{campello2021multi}
Victor~M Campello, Polyxeni Gkontra, Cristian Izquierdo, Carlos Martin-Isla, Alireza Sojoudi, Peter~M Full, Klaus Maier-Hein, Yao Zhang, Zhiqiang He, Jun Ma, et~al.
\newblock Multi-centre, multi-vendor and multi-disease cardiac segmentation: the m\&ms challenge.
\newblock {\em IEEE Transactions on Medical Imaging}, 40(12):3543--3554, 2021.

\bibitem{chang2021unsupervised}
Qi Chang, Zhennan Yan, Meng Ye, Kanski Mikael, Subhi Al’Aref, Leon Axel, and Dimitris~N Metaxas.
\newblock An unsupervised 3d recurrent neural network for slice misalignment correction in cardiac mr imaging.
\newblock In {\em International Workshop on Statistical Atlases and Computational Models of the Heart}, pages 141--150. Springer, 2021.

\bibitem{chen2018blazingly}
Yuhua Chen, Jordi Pont-Tuset, Alberto Montes, and Luc Van~Gool.
\newblock Blazingly fast video object segmentation with pixel-wise metric learning.
\newblock In {\em Proceedings of the IEEE conference on computer vision and pattern recognition}, pages 1189--1198, 2018.

\bibitem{cheng2024putting}
Ho~Kei Cheng, Seoung~Wug Oh, Brian Price, Joon-Young Lee, and Alexander Schwing.
\newblock Putting the object back into video object segmentation.
\newblock In {\em Proceedings of the IEEE/CVF Conference on Computer Vision and Pattern Recognition}, pages 3151--3161, 2024.

\bibitem{cheng2022xmem}
Ho~Kei Cheng and Alexander~G Schwing.
\newblock Xmem: Long-term video object segmentation with an atkinson-shiffrin memory model.
\newblock In {\em European Conference on Computer Vision}, pages 640--658. Springer, 2022.

\bibitem{cheng2021modular}
Ho~Kei Cheng, Yu-Wing Tai, and Chi-Keung Tang.
\newblock Modular interactive video object segmentation: Interaction-to-mask, propagation and difference-aware fusion.
\newblock In {\em Proceedings of the IEEE/CVF Conference on Computer Vision and Pattern Recognition}, pages 5559--5568, 2021.

\bibitem{cheng2021rethinking}
Ho~Kei Cheng, Yu-Wing Tai, and Chi-Keung Tang.
\newblock Rethinking space-time networks with improved memory coverage for efficient video object segmentation.
\newblock {\em Advances in Neural Information Processing Systems}, 34:11781--11794, 2021.

\bibitem{deng2009imagenet}
Jia Deng, Wei Dong, Richard Socher, Li-Jia Li, Kai Li, and Li Fei-Fei.
\newblock Imagenet: A large-scale hierarchical image database.
\newblock In {\em 2009 IEEE conference on computer vision and pattern recognition}, pages 248--255. Ieee, 2009.

\bibitem{gao2022data}
Yunhe Gao, Mu Zhou, Di Liu, Zhennan Yan, Shaoting Zhang, and Dimitris~N Metaxas.
\newblock A data-scalable transformer for medical image segmentation: architecture, model efficiency, and benchmark.
\newblock {\em arXiv preprint arXiv:2203.00131}, 2022.

\bibitem{gao2021utnet}
Yunhe Gao, Mu Zhou, and Dimitris~N Metaxas.
\newblock Utnet: a hybrid transformer architecture for medical image segmentation.
\newblock In {\em Medical Image Computing and Computer Assisted Intervention--MICCAI 2021: 24th International Conference, Strasbourg, France, September 27--October 1, 2021, Proceedings, Part III 24}, pages 61--71. Springer, 2021.

\bibitem{he2022masked}
Kaiming He, Xinlei Chen, Saining Xie, Yanghao Li, Piotr Doll{\'a}r, and Ross Girshick.
\newblock Masked autoencoders are scalable vision learners.
\newblock In {\em Proceedings of the IEEE/CVF conference on computer vision and pattern recognition}, pages 16000--16009, 2022.

\bibitem{he2016deep}
Kaiming He, Xiangyu Zhang, Shaoqing Ren, and Jian Sun.
\newblock Deep residual learning for image recognition.
\newblock In {\em Proceedings of the IEEE conference on computer vision and pattern recognition}, pages 770--778, 2016.

\bibitem{hu2017maskrnn}
Yuan-Ting Hu, Jia-Bin Huang, and Alexander Schwing.
\newblock Maskrnn: Instance level video object segmentation.
\newblock {\em Advances in neural information processing systems}, 30, 2017.

\bibitem{huttenlocher1993comparing}
Daniel~P Huttenlocher, Gregory~A. Klanderman, and William~J Rucklidge.
\newblock Comparing images using the hausdorff distance.
\newblock {\em IEEE Transactions on pattern analysis and machine intelligence}, 15(9):850--863, 1993.

\bibitem{isensee2021nnu}
Fabian Isensee, Paul~F Jaeger, Simon~AA Kohl, Jens Petersen, and Klaus~H Maier-Hein.
\newblock nnu-net: a self-configuring method for deep learning-based biomedical image segmentation.
\newblock {\em Nature methods}, 18(2):203--211, 2021.

\bibitem{khened2019fully}
Mahendra Khened, Varghese~Alex Kollerathu, and Ganapathy Krishnamurthi.
\newblock Fully convolutional multi-scale residual densenets for cardiac segmentation and automated cardiac diagnosis using ensemble of classifiers.
\newblock {\em Medical image analysis}, 51:21--45, 2019.

\bibitem{lee2009automatic}
Hae-Yeoun Lee, Noel~CF Codella, Matthew~D Cham, Jonathan~W Weinsaft, and Yi Wang.
\newblock Automatic left ventricle segmentation using iterative thresholding and an active contour model with adaptation on short-axis cardiac mri.
\newblock {\em IEEE Transactions on Biomedical Engineering}, 57(4):905--913, 2009.

\bibitem{li2010line}
Bo Li, Yingmin Liu, Christopher~J Occleshaw, Brett~R Cowan, and Alistair~A Young.
\newblock In-line automated tracking for ventricular function with magnetic resonance imaging.
\newblock {\em JACC: Cardiovascular Imaging}, 3(8):860--866, 2010.

\bibitem{li2018video}
Xiaoxiao Li and Chen~Change Loy.
\newblock Video object segmentation with joint re-identification and attention-aware mask propagation.
\newblock In {\em Proceedings of the European conference on computer vision (ECCV)}, pages 90--105, 2018.

\bibitem{liu2022isegformer}
Qin Liu, Zhenlin Xu, Yining Jiao, and Marc Niethammer.
\newblock isegformer: interactive segmentation via transformers with application to 3d knee mr images.
\newblock In {\em International Conference on Medical Image Computing and Computer-Assisted Intervention}, pages 464--474. Springer, 2022.

\bibitem{martin2023deep}
Carlos Mart{\'\i}n-Isla, V{\'\i}ctor~M Campello, Cristian Izquierdo, Kaisar Kushibar, Carla Sendra-Balcells, Polyxeni Gkontra, Alireza Sojoudi, Mitchell~J Fulton, Tewodros~Weldebirhan Arega, Kumaradevan Punithakumar, et~al.
\newblock Deep learning segmentation of the right ventricle in cardiac mri: The m\&ms challenge.
\newblock {\em IEEE Journal of Biomedical and Health Informatics}, 2023.

\bibitem{oh2018fast}
Seoung~Wug Oh, Joon-Young Lee, Kalyan Sunkavalli, and Seon~Joo Kim.
\newblock Fast video object segmentation by reference-guided mask propagation.
\newblock In {\em Proceedings of the IEEE conference on computer vision and pattern recognition}, pages 7376--7385, 2018.

\bibitem{oh2019video}
Seoung~Wug Oh, Joon-Young Lee, Ning Xu, and Seon~Joo Kim.
\newblock Video object segmentation using space-time memory networks.
\newblock In {\em Proceedings of the IEEE/CVF International Conference on Computer Vision}, pages 9226--9235, 2019.

\bibitem{perazzi2017learning}
Federico Perazzi, Anna Khoreva, Rodrigo Benenson, Bernt Schiele, and Alexander Sorkine-Hornung.
\newblock Learning video object segmentation from static images.
\newblock In {\em Proceedings of the IEEE conference on computer vision and pattern recognition}, pages 2663--2672, 2017.

\bibitem{perazzi2016benchmark}
Federico Perazzi, Jordi Pont-Tuset, Brian McWilliams, Luc Van~Gool, Markus Gross, and Alexander Sorkine-Hornung.
\newblock A benchmark dataset and evaluation methodology for video object segmentation.
\newblock In {\em Proceedings of the IEEE conference on computer vision and pattern recognition}, pages 724--732, 2016.

\bibitem{ronneberger2015u}
Olaf Ronneberger, Philipp Fischer, and Thomas Brox.
\newblock U-net: Convolutional networks for biomedical image segmentation.
\newblock In {\em International Conference on Medical image computing and computer-assisted intervention}, pages 234--241. Springer, 2015.

\bibitem{seong2021hierarchical}
Hongje Seong, Seoung~Wug Oh, Joon-Young Lee, Seongwon Lee, Suhyeon Lee, and Euntai Kim.
\newblock Hierarchical memory matching network for video object segmentation.
\newblock In {\em Proceedings of the IEEE/CVF International Conference on Computer Vision}, pages 12889--12898, 2021.

\bibitem{suinesiaputra2014collaborative}
Avan Suinesiaputra, Brett~R Cowan, Ahmed~O Al-Agamy, Mustafa~A Elattar, Nicholas Ayache, Ahmed~S Fahmy, Ayman~M Khalifa, Pau Medrano-Gracia, Marie-Pierre Jolly, Alan~H Kadish, et~al.
\newblock A collaborative resource to build consensus for automated left ventricular segmentation of cardiac mr images.
\newblock {\em Medical image analysis}, 18(1):50--62, 2014.

\bibitem{sun2023alignment}
Rui Sun, Yuan Wang, Huayu Mai, Tianzhu Zhang, and Feng Wu.
\newblock Alignment before aggregation: trajectory memory retrieval network for video object segmentation.
\newblock In {\em Proceedings of the IEEE/CVF International Conference on Computer Vision}, pages 1218--1228, 2023.

\bibitem{vaswani2017attention}
Ashish Vaswani, Noam Shazeer, Niki Parmar, Jakob Uszkoreit, Llion Jones, Aidan~N Gomez, {\L}ukasz Kaiser, and Illia Polosukhin.
\newblock Attention is all you need.
\newblock {\em Advances in neural information processing systems}, 30, 2017.

\bibitem{wang2023look}
Junke Wang, Dongdong Chen, Zuxuan Wu, Chong Luo, Chuanxin Tang, Xiyang Dai, Yucheng Zhao, Yujia Xie, Lu Yuan, and Yu-Gang Jiang.
\newblock Look before you match: Instance understanding matters in video object segmentation.
\newblock In {\em Proceedings of the IEEE/CVF Conference on Computer Vision and Pattern Recognition}, pages 2268--2278, 2023.

\bibitem{wu2023scalable}
Qiangqiang Wu, Tianyu Yang, Wei Wu, and Antoni~B Chan.
\newblock Scalable video object segmentation with simplified framework.
\newblock In {\em Proceedings of the IEEE/CVF International Conference on Computer Vision}, pages 13879--13889, 2023.

\bibitem{wu2014evaluation}
Vincent Wu, Janice~Y Chyou, Sohae Chung, Sharath Bhagavatula, and Leon Axel.
\newblock Evaluation of diastolic function by three-dimensional volume tracking of the mitral annulus with cardiovascular magnetic resonance: comparison with tissue doppler imaging.
\newblock {\em Journal of Cardiovascular Magnetic Resonance}, 16(1):1--14, 2014.

\bibitem{yang2018efficient}
Linjie Yang, Yanran Wang, Xuehan Xiong, Jianchao Yang, and Aggelos~K Katsaggelos.
\newblock Efficient video object segmentation via network modulation.
\newblock In {\em Proceedings of the IEEE conference on computer vision and pattern recognition}, pages 6499--6507, 2018.

\bibitem{yang2021associating}
Zongxin Yang, Yunchao Wei, and Yi Yang.
\newblock Associating objects with transformers for video object segmentation.
\newblock {\em Advances in Neural Information Processing Systems}, 34:2491--2502, 2021.

\bibitem{yang2022decoupling}
Zongxin Yang and Yi Yang.
\newblock Decoupling features in hierarchical propagation for video object segmentation.
\newblock {\em Advances in Neural Information Processing Systems}, 35:36324--36336, 2022.

\bibitem{ye2021deeptag}
Meng Ye, Mikael Kanski, Dong Yang, Qi Chang, Zhennan Yan, Qiaoying Huang, Leon Axel, and Dimitris Metaxas.
\newblock Deeptag: An unsupervised deep learning method for motion tracking on cardiac tagging magnetic resonance images.
\newblock In {\em Proceedings of the IEEE/CVF conference on computer vision and pattern recognition}, pages 7261--7271, 2021.

\bibitem{ye2023sequencemorph}
Meng Ye, Dong Yang, Qiaoying Huang, Mikael Kanski, Leon Axel, and Dimitris~N Metaxas.
\newblock Sequencemorph: A unified unsupervised learning framework for motion tracking on cardiac image sequences.
\newblock {\em IEEE Transactions on Pattern Analysis and Machine Intelligence}, 2023.

\bibitem{zhang2023joint}
Jiaming Zhang, Yutao Cui, Gangshan Wu, and Limin Wang.
\newblock Joint modeling of feature, correspondence, and a compressed memory for video object segmentation.
\newblock {\em arXiv preprint arXiv:2308.13505}, 2023.

\bibitem{zhang2023boosting}
Yurong Zhang, Liulei Li, Wenguan Wang, Rong Xie, Li Song, and Wenjun Zhang.
\newblock Boosting video object segmentation via space-time correspondence learning.
\newblock In {\em Proceedings of the IEEE/CVF Conference on Computer Vision and Pattern Recognition}, pages 2246--2256, 2023.

\bibitem{zhou2023nnformer}
Hong-Yu Zhou, Jiansen Guo, Yinghao Zhang, Xiaoguang Han, Lequan Yu, Liansheng Wang, and Yizhou Yu.
\newblock nnformer: Volumetric medical image segmentation via a 3d transformer.
\newblock {\em IEEE Transactions on Image Processing}, 2023.

\bibitem{zhou2023volumetric}
Tianfei Zhou, Liulei Li, Gustav Bredell, Jianwu Li, Jan Unkelbach, and Ender Konukoglu.
\newblock Volumetric memory network for interactive medical image segmentation.
\newblock {\em Medical Image Analysis}, 83:102599, 2023.

\bibitem{zhou2022survey}
Tianfei Zhou, Fatih Porikli, David~J Crandall, Luc Van~Gool, and Wenguan Wang.
\newblock A survey on deep learning technique for video segmentation.
\newblock {\em IEEE Transactions on Pattern Analysis and Machine Intelligence}, 45(6):7099--7122, 2022.

\end{thebibliography}
}

\end{document}